\documentclass{article}

\usepackage[preprint]{neurips_2025}

\usepackage[utf8]{inputenc}
\usepackage[T1]{fontenc}
\usepackage{url}
\usepackage{hyperref}
\usepackage{booktabs}
\usepackage{amsmath,amssymb,amsthm}
\usepackage{multirow}
\usepackage{graphicx}
\usepackage{xcolor}
\usepackage{enumitem}
\usepackage{algorithm}
\usepackage{algpseudocode}
\usepackage{caption}
\usepackage{subcaption}
\usepackage{array}
\usepackage{tabularx}
\usepackage{colortbl}
\usepackage{setspace}

\usepackage{listings}
\usepackage{xcolor}

\usepackage[normalem]{ulem}
\usepackage{float}

\definecolor{agentblue}{RGB}{79, 70, 229}
\definecolor{lightgray}{RGB}{245, 245, 245}
\definecolor{goodgreen}{RGB}{22, 163, 74}
\definecolor{badred}{RGB}{220, 38, 38}
\definecolor{highlightrow}{RGB}{236, 252, 243}

\hypersetup{
  colorlinks=true,
  linkcolor=agentblue,
  citecolor=agentblue,
  urlcolor=agentblue,
}

\newcommand{\combo}{\mathbf{c}}
\newcommand{\combospace}{\mathcal{C}}

\theoremstyle{definition}

\title{\textbf{AgentOpt v0.1 Technical Report:\\ Client-Side Optimization for LLM-Based Agent}}

\author{%
  \textbf{Wenyue Hua}$^{1*}$, \textbf{Sripad Karne}$^{3}$, \textbf{Qian Xie}$^{2}$,  \textbf{Armaan Agrawal}$^{3}$,
  \textbf{Nikos Pagonas}$^{3}$,\\ \textbf{Kostis Kaffes}$^{3}$, \textbf{Tianyi Peng}$^{3*}\thanks{Correspondence: wenyuehua@microsoft.com, kkaffes@cs.columbia.edu, tianyi.peng@columbia.edu}$\\
  $^{1}$ Microsoft Research, AI Frontiers \qquad
  $^{2}$ Cornell University \qquad
  $^{3}$ Columbia University \\
  $^{*}$ Equal contribution
}

\begin{document}
\maketitle

\begin{abstract}
AI agents are increasingly deployed in real-world applications, including systems such as Manus, OpenClaw, and coding agents. Existing research has primarily focused on \emph{server-side} efficiency, proposing methods such as caching, speculative execution, traffic scheduling, and load balancing to reduce the cost of serving agentic workloads. However, as users increasingly construct agents by composing local tools, remote APIs, and diverse models, an equally important optimization problem arises on the \emph{client side}. Client-side optimization asks how developers should allocate the resources available to them, including model choice, local tools, and API budget across pipeline stages, subject to application-specific quality, cost, and latency constraints. Because these objectives depend on the task and deployment setting, they cannot be determined by server-side systems alone. We introduce \textsc{AgentOpt}, the first framework-agnostic Python package for client-side agent optimization. We first study \emph{model selection}, a high-impact optimization lever in multi-step agent pipelines. Given a pipeline and a small evaluation set, the goal is to find the most cost-effective assignment of models to pipeline roles. This problem is consequential in practice: at matched accuracy, the cost gap between the best and worst model combinations can reach 13--32$\times$ in our experiments. To efficiently explore the exponentially growing combination space, \textsc{AgentOpt} implements ten search algorithms, including UCB-E, UCB-E with Low-Rank Factorization, Arm Elimination, Epsilon-LUCB, Threshold Successive Elimination, and Bayesian Optimization. Across four benchmarks, UCB-E recovers near-optimal accuracy while reducing evaluation budget by 62-76\% relative to brute-force search.

\medskip
\noindent\textbf{Code and benchmarks:} \url{https://agentoptimizer.github.io/agentopt/}

\end{abstract}

\section{Introduction}
\label{sec:intro}

AI agents have rapidly evolved from research prototypes into widely deployed systems. Open-source frameworks such as Langgraph~\citep{wang2024agent}, AutoGen~\citep{wu2024autogen}, and OpenClaw~\citep{shan2026don, wang2026openclaw}, as well as commercial systems such as Manus and ClaudeCode~\citep{chatlatanagulchai2025use}, are increasingly used in research and practice. This adoption has motivated substantial interest in improving the efficiency of agent execution. Recent work has largely approached this problem from the \emph{server side}, developing techniques such as request scheduling~\citep{ni2026chimera, song2025gradientsys, li2025throughput}, load balancing~\citep{coviello2025bifrost, deng2025agentic}, speculative execution~\citep{hua2024interactive, guan2025dynamic, ye2025speculative}, and cache reuse~\citep{chillara2026semanticalli, begum2026hierarchical, zhang2025agentic} to reduce the cost and latency of serving agentic workloads. Systems such as \textbf{Autellix}~\citep{luo2025autellix}, \textbf{ThunderAgent}~\citep{kang2026thunderagent}, \textbf{Continuum}~\citep{li2025continuum}, and \textbf{AIOS}~\citep{mei2024aios} exemplify this trend. Compared with traditional LLM serving systems such as vLLM~\citep{kwon2023vllm} and SGLang~\citep{zheng2024sglang}, which optimize individual model calls, these agent-serving systems target end-to-end execution over multi-step reasoning and tool-use trajectories.

Despite this progress, server-side optimization captures only part of the efficiency problem. Increasingly, users are not merely API consumers but agent builders: they compose pipelines from heterogeneous components, including local tools, remote APIs, and multiple candidate models. In this setting, many critical resources and decisions lie on the \emph{client side}. Developers choose which model to assign to each role, how to allocate API budget across steps, when to invoke local tools, and which quality-cost-latency tradeoff is acceptable for a particular application. These decisions depend on task requirements, budget constraints, and service-level objectives that are specific to the deployment setting and are often unavailable to the model provider. As a result, client-side optimization exposes an efficiency surface that is fundamentally different from the one addressed by server-side systems.

This distinction is practically important. A provider may optimize for throughput, tail latency, or cluster utilization across many users, whereas an individual developer may care about a very different objective. A startup building a coding assistant may accept a modest drop in accuracy for a large reduction in cost, while a high-stakes clinical decision-support system may prioritize reliability over all other considerations. Such choices cannot be determined centrally, because they are inseparable from application-specific utility. Client-side optimization is therefore the layer at which a developer's own objectives can be directly and transparently optimized.

Among the optimization levers available on the client side, we argue that \emph{model selection} has the largest empirical impact. Choices such as caching, routing, and scheduling matter only after a model assignment has been fixed. By contrast, selecting the wrong model combination can dominate all downstream efficiency improvements. In our experiments, the cost gap between the best and worst model combinations at matched accuracy ranges from 13$\times$ to 32$\times$ across benchmarks. On BFCL, for example, Qwen3 Next 80B matches Claude Opus 4.6 in accuracy while reducing cost by 32$\times$. Gaps of this magnitude cannot be closed by serving optimizations alone.

At a high level, our problem can be viewed as LLM routing~\citep{hu2024routerbench, mei2025omnirouter, zhang2025leveraging} in an end-to-end agent pipeline, but it is fundamentally different from standard routing. In conventional LLM routing, each query is assigned to a stronger or cheaper model based on its estimated difficulty, and decisions are typically made at the level of individual calls. In multi-step agent pipelines, by contrast, routing decisions are coupled across stages. When multiple candidate models are available for multiple pipeline roles, the search space grows combinatorially, and the utility of any local assignment depends on its downstream effect on the overall trajectory. For example, a planner agent should be evaluated not in isolation, but by whether its output enables the solver agent and tool calls to proceed effectively. Agent model routing is therefore better understood as an end-to-end sequential decision problem, naturally formulated as a Markov decision process, in which each routing choice influences future states and final utility through cross-step interactions. As a result, optimizing each step independently is generally insufficient; effective optimization must operate over full model combinations or pipeline-level routing policies.

Our empirical results illustrate the huge impact of model selection in agentic pipeline. On HotpotQA~\citep{yang2018hotpotqa}, Claude Opus 4.6, the strongest model in our benchmark by standalone capability, is the worst planner across all 81 model combinations. When used as planner, it often answers directly from parametric knowledge and bypasses the solver's search tools, preventing the downstream system from executing the intended reasoning process. By contrast, Ministral 3 8B, the cheapest planner in the benchmark, more reliably delegates to the solver. Paired with Opus as solver, this combination achieves 74.27\% accuracy, whereas using Opus for both roles yields only 31.71\%. The key point is that neither model can be judged in isolation: the relevant object of evaluation is the \emph{combination}. Agent model selection is therefore a combination-level optimization problem rather than a sequence of independent routing decisions.

\begin{figure}[t]
\centering
\includegraphics[width=\textwidth]{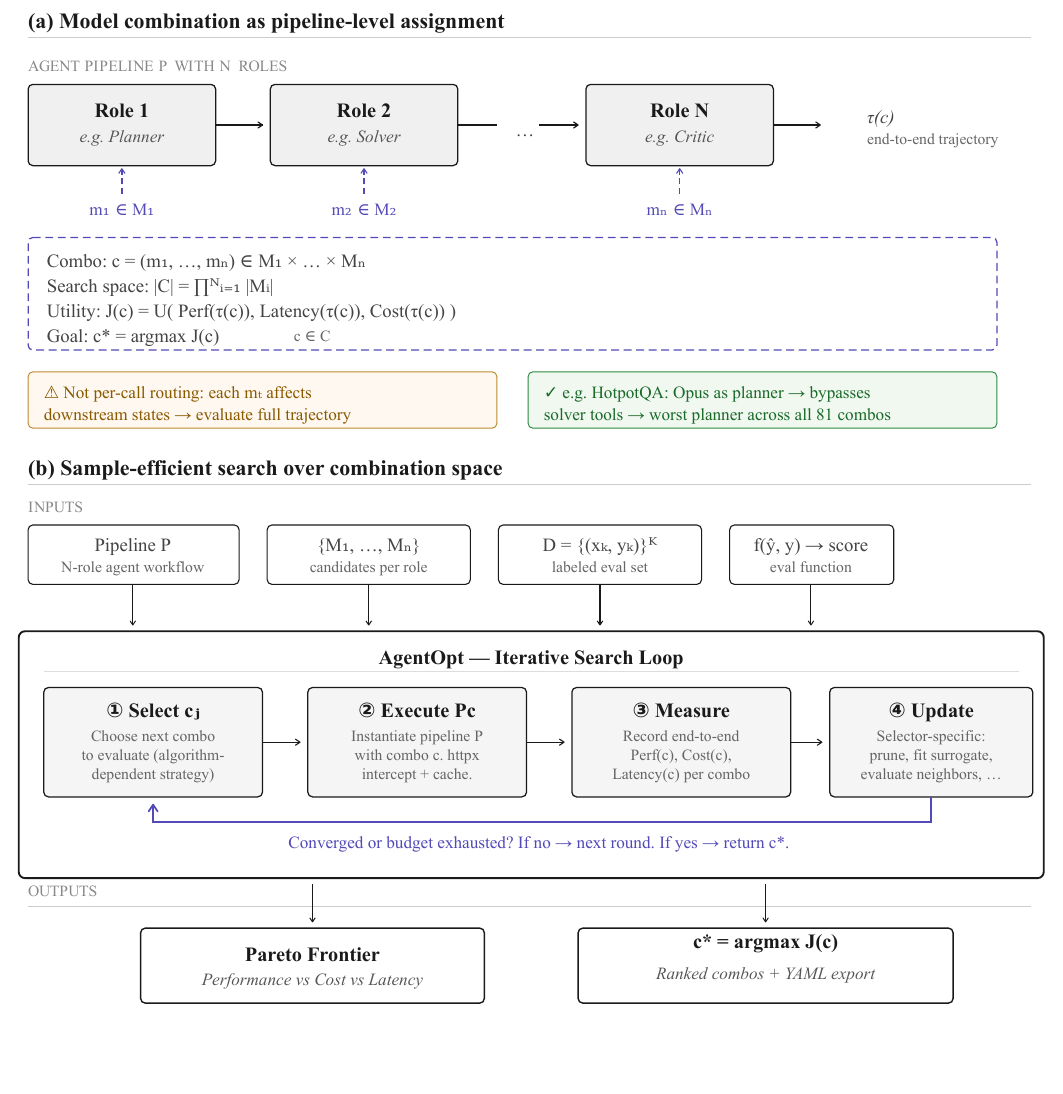}
\caption{Overview of \textsc{AgentOpt}. \textbf{(a)} A model combination $c = (m_1, \dots, m_N)$ assigns one model per pipeline role. The full trajectory $\tau(c)$ is evaluated end-to-end. \textbf{(b)} The optimization loop iteratively selects, executes, and measures combinations, returning the Pareto frontier over performance, cost, and latency.}
\label{fig:system-overview}
\end{figure}

To address this problem, we present \textsc{AgentOpt}, \textbf{an open-source Python package for client-side optimization} and the first functionality supported is model combination selection for agentic workflows. \textsc{AgentOpt} is framework-agnostic and operates by intercepting LLM calls at the HTTP transport layer, allowing it to support existing agent implementations with minimal integration overhead. Given an agent pipeline, a candidate model pool, and a labeled evaluation set, \textsc{AgentOpt} searches the space of model assignments and returns the Pareto frontier over accuracy, cost, and latency. To make this search practical, it implements ten algorithms, including multi-armed bandit methods and Bayesian optimization, for efficient exploration of a combinatorial space that grows exponentially with the number of pipeline roles (Figure~\ref{fig:system-overview}).

Our contributions are threefold. First, we formalize \emph{client-side optimization} as a distinct systems problem for AI agents and identify model combination selection as a first-class optimization target. Second, we show why multi-step agent optimization must be performed at the level of full model combinations rather than individual calls, and introduce the \emph{combo abstraction} as the appropriate unit of evaluation. Third, we implement and evaluate ten search algorithms for sample-efficient exploration of the combination space, showing that matrix-based upper confidence bound exploration algorithm (Matrix UCB-E)~\citep{zhou2025speeding} recovers near-optimal combinations while reducing evaluation cost by 62--76\% relative to brute-force search across all four benchmarks.

The remainder of the technical report is organized as follows. Section~\ref{sec:user-side} defines the client-side optimization problem and the associated tradeoff space. Section~\ref{sec:model-selection} motivates model selection as the dominant optimization lever. Section~\ref{sec:agent-routing} distinguishes agent model selection from standard LLM routing and introduces the combo abstraction. Section~\ref{sec:overhead} presents the search algorithms used to make combination-level evaluation tractable. Section~\ref{sec:empirics} reports empirical results across four benchmarks. Section~\ref{sec:related} reviews related work, and Section~\ref{sec:conclusion} concludes.

\section{Client-Side Optimization}
\label{sec:user-side}

Recent systems work on agent efficiency has focused primarily on the \emph{server side}: improving the deployment of agentic workloads through scheduling, caching, speculative execution, and load balancing. These techniques are important, but they optimize the provider's infrastructure rather than the developer's application. As AI agents become increasingly customizable, a complementary optimization problem emerges on the \emph{client side}, where the agent developer decides how to assemble and run a pipeline using the resources directly available to them.

Client-side optimization refers to improving an agentic workflow using resources and decisions under the user's control. These include the set of candidate foundation models accessible through APIs or local deployment, the assignment of models to different roles in a pipeline, the invocation of local and remote tools, the allocation of API budget across steps, and operational choices such as batching, caching, and scheduling within the application itself. Unlike server-side systems, which optimize shared infrastructure across many users, client-side optimization operates at the level of a specific workflow and a specific utility function. It is therefore the natural layer for optimizing an agent according to the developer's own requirements.

This perspective is increasingly important because modern agent pipelines are no longer monolithic. Developers routinely compose planners, solvers, critics, retrievers, and external tools into multi-stage workflows, and many of the relevant resources now sit outside provider control. Some resources are financial, such as per-query API budget; some are computational, such as local model execution or tool latency; and some are structural, such as which tools or sub-agents are available in the workflow. The central question is therefore no longer only how providers can serve agents efficiently, but how users can configure their own agent systems efficiently given the resources they actually possess.

Client-side optimization also differs from server-side optimization in a second and more fundamental way: the objective is inherently personalized. The utility of an agent depends on application-specific preferences over quality, latency, and monetary cost, and these preferences vary widely across deployments. A coding assistant used in an interactive setting may accept some loss in task performance in exchange for lower latency or lower API spend. A medical or legal assistant may instead prioritize quality almost exclusively. These choices depend on the developer's task, budget, and service requirements, and cannot be inferred by the provider from system-level signals alone. Client-side optimization is therefore the layer at which application-specific objectives can be expressed directly.

We formalize this setting through a general performance-latency-cost tradeoff. Consider an agent pipeline with $N$ roles and a candidate model set $\mathcal{M}$. A \emph{model combination} is an assignment $\combo = (m_1,\dots,m_N) \in \mathcal{M}^N$ specifying which model is used for each role. More broadly, one may view a client-side configuration as specifying not only model assignment but also other controllable decisions such as tool routing or scheduling policy. Each such configuration induces three quantities of practical interest: task performance, execution latency, and monetary cost. The developer's goal is not to optimize any one of these in isolation, but to identify configurations that best match the desired tradeoff among them.

Formally, let $\mathcal{D} = \{(x_k, y_k)\}_{k=1}^K$ denote a labeled evaluation set, and let $P_\theta$ denote the pipeline instantiated under a client-side configuration $\theta$. We write
\[
\textsc{Perf}(\theta), \qquad \textsc{Latency}(\theta), \qquad \textsc{Cost}(\theta)
\]
for the resulting end-to-end task performance, latency, and cost. The developer then seeks configurations on the Pareto frontier, or equivalently optimizes a user-defined utility
\[
U(\theta) = U\big(\textsc{Perf}(\theta), \textsc{Latency}(\theta), \textsc{Cost}(\theta)\big),
\]
possibly under explicit constraints on latency or budget. This formulation is intentionally general: different applications may use exact-match accuracy, pass rate, or LLM-as-judge metrics for performance, while weighting latency and cost according to deployment-specific requirements.

\section{Model Selection in Agent}
\label{sec:model-selection}

Among the many levers available on the client side, we argue that \emph{model selection} is the first-class optimization problem. The reason is structural. Most other optimizations available to developers, such as caching, request scheduling, speculative execution, or tool-level heuristics, operate conditional on a model assignment. They improve the efficiency of a pipeline once the participating models have already been chosen. Model selection instead determines the computational substrate on which all subsequent optimizations act. In this sense, it is upstream of the rest.

This distinction is not merely conceptual; it is empirically large. Across our benchmarks, the gap between the best and worst model combinations at comparable task quality ranges from 13$\times$ to 32$\times$ in cost. On BFCL, Qwen3 Next 80B matches Claude Opus 4.6 in accuracy at 32$\times$ lower cost. On MathQA, the gap between expensive and budget-efficient combinations reaches 24$\times$ while maintaining similar accuracy. These differences are too large to be recovered by infrastructure improvements alone. A pipeline built on the wrong model assignment can be dominated by one built on the right assignment even before any additional systems optimization is applied.

A second reason model selection must be treated as first-class is that performance rankings do not transfer cleanly across roles. In multi-step agents, the quality of a model is not a context-free property; it depends on how that model behaves in a particular role and how its behavior interacts with downstream components. A strong standalone model may be an excellent solver yet a poor planner, or vice versa. Our HotpotQA experiments illustrate this clearly: Claude Opus 4.6 is the strongest model in absolute capability, yet performs worst as a planner in our two-stage setup, while a much smaller model, Ministral 3 8B, is the best planner because it more reliably delegates to the downstream solver and tool chain. The resulting best-performing combination pairs a weak planner with a strong solver, demonstrating that model choice must be evaluated at the level of the workflow rather than the model in isolation.

These results motivate a shift in viewpoint. The relevant optimization object is not the ``best model'' in the abstract, nor even the best model for each role considered independently, but the best \emph{combination} of models for a particular agentic workflow under a particular utility function. Model selection in agents is therefore a combinatorial optimization problem over pipeline-level assignments.

\subsection{Model Selection in Agentic Workflows as Black-Box Optimization}
\label{sec:agent-routing}

At a high level, this problem can be viewed as LLM routing for agentic pipelines. However, its mathematical structure differs substantially from standard single-call routing. In conventional LLM routing, a router selects among candidate models for an individual query based on estimated difficulty or expected utility. The optimization unit is a single model call. In multi-step agentic workflows, by contrast, model choices interact across stages: assigning a model to one role changes the intermediate computation encountered by later roles, and the value of any local assignment depends on its downstream effect on the overall trajectory.

This dependence makes agent model selection naturally a \emph{black-box optimization problem}~\citep{kumagai2023black, golovin2017google} over full pipeline configurations. Consider a pipeline with $H$ roles and a candidate model set $M$. A model combination is a tuple
\[
\combo = (m_1, \dots, m_H) \in M^H,
\]
where $m_t$ denotes the model assigned to role $t$. Given an input query, executing the pipeline under combination $\combo$ induces an end-to-end trajectory $\tau(\combo)$, including all intermediate model outputs, tool interactions, latency, and monetary cost. Since these cross-stage interactions are generally complex and task-dependent, the resulting utility is most naturally treated as an unknown black-box function of the full combination.

A convenient formulation is
\[
J(c) = U\!\big(\mathrm{PERF}(\tau(\combo)), \mathrm{LATENCY}(\tau(\combo)), \mathrm{COST}(\tau(\combo))\big),
\]
where $U(\cdot)$ is a user-specified utility function. Equivalently, one may write
\[
J(\combo) = \mathrm{PERF}(\tau(\combo)) - \lambda_c\,\mathrm{COST}(\tau(\combo)) - \lambda_\ell\,\mathrm{LATENCY}(\tau(\combo)),
\]
for task-dependent weights $\lambda_{\combo}, \lambda_\ell \ge 0$. The client-side optimization problem then becomes
\[
\combo^\star = \arg\max_{\combo \in M^H} J(\combo).
\]

Under this view, the key consequence is that the value of assigning a model at stage $t$ cannot in general be evaluated independently of the downstream assignments, because it affects later computation, future tool usage, and the final end-to-end utility. As a result, optimizing each step independently is generally insufficient; effective optimization must operate over full model combinations or pipeline-level routing strategies.
\section{AgentOpt Package Design}
\label{sec:system}

The formulation above defines client-side model selection as an end-to-end optimization problem over agentic workflows. To make this problem practical for real systems, we build \textsc{AgentOpt}, a framework-agnostic Python package that exposes model-combination optimization through a simple evaluation API while transparently handling LLM interception, metric collection, caching, and parallel execution.

\subsection{Design Goals}

\textsc{AgentOpt} is designed around three requirements. First, it should be \emph{framework-agnostic}: developers should be able to optimize agents written using different LLM libraries and orchestration frameworks without rewriting the agent itself. Second, it should be \emph{non-intrusive}: optimization should not require proxy servers, custom wrappers around each SDK, or manual instrumentation of every LLM call. Third, it should provide a unified \emph{execution substrate} for model selection, so that different search algorithms can operate on the same measured quantities of task performance, latency, and cost.

These goals motivate a separation of concerns between \emph{selection policy} and \emph{execution infrastructure}. The selector decides which model combinations to evaluate; the package runtime is responsible for executing those combinations, attributing the resulting LLM calls to the correct datapoints and combinations, and aggregating end-to-end statistics. This separation is important for the rest of the paper: the algorithms in Section~\ref{sec:overhead} differ in how they explore the combination space, but they all rely on the same system substrate for evaluation.

\subsection{User-Facing API}

At the API level, \textsc{AgentOpt} treats model selection as optimization over an existing agent implementation rather than as a new programming framework. The developer supplies an agent class, a candidate model set for each role, a labeled evaluation dataset, and a task-specific evaluation function. A selector then searches over model combinations and returns a structured result object containing ranked combinations and their associated metrics.

Concretely, the agent interface is intentionally minimal: the package expects an agent class with a constructor that accepts a model assignment and a \texttt{run()} method that executes the workflow on a datapoint. Candidate models are provided as a dictionary mapping role names to model lists. The evaluation function maps an expected output and an actual output to a score, allowing developers to use exact match, task-specific metrics, or LLM-as-judge scoring depending on the application. The same interface is shared across all selectors, which enables direct substitution of brute-force search, bandit methods, hill climbing, Bayesian optimization, and other search strategies without changing the surrounding code.

A typical usage pattern is:

\begin{lstlisting}[language=Python, caption={Model selection main API.}, label={lst:main_api}]
selector = ArmEliminationModelSelector(
    agent=MyAgent,
    models={
        "planner": [...],
        "solver": [...],
        "critic": [...],
    },
    eval_fn=eval_fn,
    dataset=dataset,
    model_prices=model_prices,
)

results = selector.select_best(parallel=True, max_concurrent=20)
\end{lstlisting}

The output is a \texttt{SelectionResults} object that exposes both human-readable summaries and programmatic access to the optimized configuration. In particular, the package can print ranked tables, return the top-performing combination, export all evaluated results to CSV, and export the selected configuration as a YAML file for downstream deployment. This design makes \textsc{AgentOpt} usable both as an interactive development tool and as part of a reproducible optimization pipeline.

\subsection{Framework-Agnostic Interception and Attribution}

The main systems challenge is to observe and control LLM calls across diverse agent frameworks without requiring framework-specific adapters. \textsc{AgentOpt} addresses this by intercepting requests at the HTTP transport layer, patching \texttt{httpx.Client.send()} and \texttt{httpx.AsyncClient.send()} so that LLM calls can be tracked uniformly regardless of the upstream framework. Attribution is handled with Python \texttt{contextvars}, which associates each intercepted call with the current datapoint and model combination.

Listing~\ref{lst:http-intercept} shows a simplified version of the mechanism.

\begin{lstlisting}[language=Python, caption={Simplified HTTP-layer interception and attribution in \textsc{AgentOpt}.}, label={lst:http-intercept}]
import contextvars
import httpx
from agentopt.proxy import LLMTracker

tracker = LLMTracker(cache=True)

_current_data_id = contextvars.ContextVar("data_id", default=None)
_current_combo_id = contextvars.ContextVar("combo_id", default=None)

_original_send = httpx.Client.send
_original_async_send = httpx.AsyncClient.send

def patched_send(self, request, *args, **kwargs):
    data_id = _current_data_id.get()
    combo_id = _current_combo_id.get()

    with tracker.track(data_id=data_id, combo_id=combo_id):
        return _original_send(self, request, *args, **kwargs)

async def patched_async_send(self, request, *args, **kwargs):
    data_id = _current_data_id.get()
    combo_id = _current_combo_id.get()

    with tracker.track(data_id=data_id, combo_id=combo_id):
        return await _original_async_send(self, request, *args, **kwargs)

httpx.Client.send = patched_send
httpx.AsyncClient.send = patched_async_send
\end{lstlisting}

This transport-layer design provides two benefits. First, it makes the package broadly compatible with heterogeneous developer stacks. Second, it allows \textsc{AgentOpt} to collect metrics uniformly across frameworks. For each intercepted call, the runtime records the model name, input and output token counts, wall-clock latency, whether the response was served from cache, and metadata identifying which datapoint and which model combination triggered the call. These low-level call records are then aggregated into per-datapoint and per-combination statistics that form the basis of the end-to-end optimization objective.

In addition, agent code remains unchanged. Any framework that ultimately issues LLM requests through \texttt{httpx} is automatically covered by the interception layer, while the tracker records model name, token usage, latency, cache status, and the datapoint/combination identifiers needed for end-to-end evaluation. 

\subsection{Caching and Parallel Evaluation}

Because model selection requires repeated execution of similar workflows, the runtime includes two mechanisms to reduce wall-clock overhead and redundant API spend: HTTP-level response caching and bounded parallel evaluation.

The caching layer stores responses keyed by a hash of the request payload, so identical LLM calls arising across different combinations or repeated runs are reused rather than re-issued. This is particularly valuable in agent pipelines where different model combinations often share some calls exactly. For example, two combinations that use the same planner model will typically induce identical planner requests on the same datapoint. \textsc{AgentOpt} therefore maintains an in-memory cache during execution and optionally persists cache entries to SQLite on disk, enabling cheap re-runs and recovery from interrupted optimization jobs. Importantly, cached entries retain their original latency measurements, so that downstream latency comparisons remain faithful to the original execution rather than being artificially biased by cache hits.

Parallel execution is controlled through the \texttt{parallel=True} and \texttt{max\_concurrent} arguments to \texttt{select\_best()}. Rather than launching all combinations and all datapoints simultaneously, which would quickly exceed API rate limits, \textsc{AgentOpt} uses a two-level concurrency scheme. One semaphore controls how many combinations are evaluated at once, and another controls how many datapoints are processed concurrently within each combination. This design preserves a global bound on in-flight API calls while still exploiting available parallelism. It also adapts naturally to different search strategies: full-dataset methods such as brute-force search benefit from datapoint-level parallelism within each combination, whereas bandit-style methods that evaluate small batches benefit more from running many combinations concurrently.

\subsection{Results, Export, and Role in the Optimization Stack}

The final output of a selector run is not just a single winning combination, but a structured record of the explored design space. Each evaluated combination is represented by a result object containing aggregate accuracy, latency, token usage, and estimated price, together with a pareto-frontier diagram. This structure is useful for more than ranking alone. It enables a transparent Pareto analysis, cost-quality tradeoff inspection, and export of optimized configurations into downstream applications.

From the perspective of the overall package design, this output layer completes the abstraction boundary. The execution substrate intercepts and measures LLM calls, the selector determines which combinations to explore, and the results object exposes the optimized frontier back to the developer. This separation lets the package support multiple search algorithms without changing the user-facing workflow. In the next section, we build on this substrate and study how to explore the exponentially large combination space efficiently.
\section{Sample-Efficient Search for Model Selection}
\label{sec:overhead}

The main practical challenge in client-side model selection is the cost of evaluating the black-box objective introduced in Section~3.1. Recall that each model combination $\combo \in C$ corresponds to a full pipeline configuration, and its utility $J(\combo)$ can only be estimated by executing the agent end-to-end on a labeled dataset. Let a pipeline contain $N$ roles, and let $M_i$ denote the candidate model set for role $i$. The resulting search space is
\[
C = \prod_{i=1}^N M_i,
\]
with cardinality
\[
|C| = \prod_{i=1}^N |M_i|.
\]
When all roles share the same candidate pool $M$, this reduces to $|C| = |M|^N$. Evaluating a single combination requires running the full pipeline across the dataset, making the total cost of exhaustive search scale as $O(K|C|)$, where $K$ is the dataset size. This quickly becomes prohibitive even for moderate pipeline sizes.

The goal of sample-efficient search is therefore to identify a near-optimal combination while minimizing the number of full pipeline evaluations. Different algorithms in AGENTOPT approach this problem with different assumptions about the structure of $J(\combo)$: some treat combinations as independent candidates and adaptively allocate samples across them, while others exploit structure in the search space or fit surrogate models to guide exploration.

\subsection{Bandit Formulation for Adaptive Elimination Methods}

For elimination-style methods, model selection can be viewed as a pure-exploration multi-armed bandit problem~\citep{slivkins2019introduction, vermorel2005multi, kuleshov2014algorithms}. Each combination $\combo_j \in \combospace$ is an arm. Pulling arm $\combo_j$ once means evaluating $P_{\combo_j}$ on one datapoint $(x_k,y_k)$ and observing reward
\[
r_{j,k} = f\!\left(P_{\combo_j}(x_k), y_k\right).
\]
After $n_j$ pulls, the empirical mean is
\[
\hat{J}_j = \frac{1}{n_j}\sum_{k=1}^{n_j} r_{j,k}.
\]
Adaptive search methods then allocate additional evaluations based on the uncertainty of each $\hat{J}_j$, seeking either the best arm, an $\varepsilon$-optimal arm, or all arms above a quality threshold.

This bandit view motivates our use of three bandit algorithms: Arm Elimination, Epsilon-LUCB~\citep{garivier2011upper, garivier2008upper}, and Threshold Successive Elimination~\citep{huang2006threshold}. It does not cover all search strategies in the package, for which we also consider two additional black-box algorithms. Hill Climbing~\citep{chinnasamy2022review, selman2006hill} instead assumes a neighborhood structure over combinations, while Bayesian Optimization~\citep{frazier2018tutorial} fits a surrogate model over the search space.

\subsection{Algorithms}

\paragraph{Brute-force search.}
Brute-force search evaluates every combination on the full dataset and returns the exact optimum within the candidate pool. It serves as the gold-standard baseline but scales poorly with search-space size.

\paragraph{Random search.}
Random search samples a subset of combinations uniformly at random and evaluates only those combinations. It is a simple baseline for budget-limited settings.

\paragraph{Matrix UCB-E.}
Matrix UCB-E treats the combination-by-datapoint evaluation grid as a matrix and selects combinations using Upper Confidence Bound scoring: $\text{UCB}_i = \bar{s}_i + \sqrt{a / n_i}$, where $\bar{s}_i$ is the mean observed score and $n_i$ is the observation count. Each step evaluates a batch of unobserved datapoints for the highest-UCB combination. An observation budget fraction controls what proportion of the grid is observed before stopping.
\begin{algorithm}[!ht]
\caption{Matrix UCB-E for model-combination search}
\label{alg:matrix-ucb}
\begin{algorithmic}[1]
\Require Combination set $\combospace$, dataset $\mathcal{D}$, exploration weight $a$, batch size $B$, budget fraction $\beta$
\State Initialize score matrix $S \in \mathbb{R}^{|\combospace| \times |\mathcal{D}|}$ with all entries unobserved
\State $N \gets \lceil \beta \cdot |\combospace| \cdot |\mathcal{D}| \rceil$ \Comment{Total observation budget}
\State $n_{\text{obs}} \gets 0$
\While{$n_{\text{obs}} < N$}
    \ForAll{$\combo_j \in \combospace$}
        \If{$\combo_j$ has observations}
            \State $\text{UCB}_j \gets \hat{\mu}_j + \sqrt{a \,/\, n_j}$
        \Else
            \State $\text{UCB}_j \gets \infty$ \Comment{Unvisited combos explored first}
        \EndIf
        \If{$\combo_j$ fully observed}
            \State $\text{UCB}_j \gets -\infty$
        \EndIf
    \EndFor
    \State $j^\star \gets \arg\max_j \text{UCB}_j$
    \State Sample $k = \min(B,\; N - n_{\text{obs}})$ unobserved datapoints for $\combo_{j^\star}$
    \State Evaluate $\combo_{j^\star}$ on selected datapoints; record scores in $S$
    \State $n_{\text{obs}} \gets n_{\text{obs}} + k$
\EndWhile
\State \Return $\arg\max_{\combo_j} \hat{\mu}_j$ among observed combinations
\end{algorithmic}
\end{algorithm}

\paragraph{Matrix UCB-E-LRF.}
Matrix UCB-E-LRF extends Matrix UCB-E by fitting a low-rank approximation $X \approx UV^\top$ to the partially observed score matrix. An ensemble of factorizations with random dropout provides uncertainty estimates, targeting high-uncertainty cells for observation after an initial random warmup phase. UCB-E-LRF is designed to improve sample efficiency on harder problems where performance differences between combinations are subtle, by exploiting correlation structure in the scoring matrix to predict unobserved entries and guide exploration more effectively~\citep{zhou2025speeding}.

\begin{algorithm}[!ht]
\caption{Matrix UCB-E-LRF for model-combination search}
\label{alg:matrix-ucb-lrf}
\begin{algorithmic}[1]
\Require Combination set $\combospace$, dataset $\mathcal{D}$, rank $r$, ensemble size $E$, warmup fraction $w$, uncertainty weight $\eta$, batch size $B$, budget fraction $\beta$
\State Initialize score matrix $S \in \mathbb{R}^{|\combospace| \times |\mathcal{D}|}$ with all entries unobserved
\State $N \gets \lceil \beta \cdot |\combospace| \cdot |\mathcal{D}| \rceil$ \Comment{Total observation budget}
\State $n_{\text{obs}} \gets 0$
\While{$n_{\text{obs}} < N$}
    \If{$n_{\text{obs}} / N_{\text{available}} < w$} \Comment{Warmup phase}
        \State Sample $k = \min(B,\; N - n_{\text{obs}})$ random unobserved cells
        \State Evaluate and record scores in $S$
        \State $n_{\text{obs}} \gets n_{\text{obs}} + k$
    \Else \Comment{LRF-guided phase}
        \State Fit ensemble of $E$ low-rank factorizations $\{S \approx U_e V_e^\top\}_{e=1}^{E}$ via ALS with dropout
        \State $\hat{\mu}_{ij} \gets \frac{1}{E} \sum_{e=1}^{E} (U_e V_e^\top)_{ij}$ \Comment{Mean prediction}
        \State $\hat{\sigma}_{ij} \gets \text{std}_{e=1}^{E} (U_e V_e^\top)_{ij}$ \Comment{Uncertainty}
        \State For observed cells: $\text{UCB}_{ij} \gets S_{ij}$; for unobserved: $\text{UCB}_{ij} \gets \hat{\mu}_{ij} + \eta \cdot \hat{\sigma}_{ij}$
        \State $\text{UCB}_j \gets \frac{1}{|\mathcal{D}|} \sum_i \text{UCB}_{ji}$ for each combination $j$
        \State $j^\star \gets \arg\max_j \text{UCB}_j$
        \State Select $k = \min(B,\; N - n_{\text{obs}})$ unobserved datapoints for $\combo_{j^\star}$ with highest $\hat{\sigma}_{j^\star, i}$
        \State Evaluate and record scores in $S$
        \State $n_{\text{obs}} \gets n_{\text{obs}} + k$
    \EndIf
\EndWhile
\State \Return $\arg\max_{\combo_j} \hat{\mu}_j$ among observed combinations
\end{algorithmic}
\end{algorithm}

\paragraph{Arm Elimination.}
Arm Elimination is the default adaptive method in \textsc{AgentOpt}. It proceeds in rounds, evaluates surviving combinations on progressively larger batches, and removes combinations whose upper confidence bounds fall below the lower confidence bounds of current leaders. This design is effective when the quality distribution is skewed and many combinations can be ruled out after a small number of evaluations.

\begin{algorithm}[!ht]
\caption{Arm Elimination for model-combination search}
\label{alg:arm-elim}
\begin{algorithmic}[1]
\Require Combination set $\combospace$, dataset $\mathcal{D}$, batch schedule $\{b_t\}_{t=1}^T$
\State $\mathcal{A}_1 \gets \combospace$
\For{$t = 1,2,\dots,T$}
    \ForAll{$\combo_j \in \mathcal{A}_t$}
        \State Evaluate $\combo_j$ on the next batch of $b_t$ datapoints
        \State Update empirical mean $\hat{J}_j$ and confidence interval $[L_j,U_j]$
    \EndFor
    \State Let $L^\star \gets \max_{\combo_j \in \mathcal{A}_t} L_j$
    \State Eliminate all $\combo_j$ such that $U_j < L^\star$
    \State $\mathcal{A}_{t+1} \gets \{\combo_j \in \mathcal{A}_t : U_j \ge L^\star\}$
    \If{$|\mathcal{A}_{t+1}| = 1$}
        \State \Return the remaining combination
    \EndIf
\EndFor
\State \Return $\arg\max_{\combo_j \in \mathcal{A}_{T+1}} \hat{J}_j$
\end{algorithmic}
\end{algorithm}

\paragraph{Epsilon-LUCB.}
Epsilon-LUCB targets $\varepsilon$-optimal best-arm identification. At each round, it maintains the empirical leader and its strongest challenger under upper confidence bounds, and allocates additional samples to the most ambiguous pair. It stops once the leader is separated from the challenger by at least $\varepsilon$.

\begin{algorithm}[!ht]
\caption{Epsilon-LUCB for model-combination search}
\label{alg:eps-lucb}
\begin{algorithmic}[1]
\Require Combination set $\combospace$, tolerance $\varepsilon$
\State Initialize all combinations with a small number of evaluations
\Repeat
    \State Compute empirical means $\hat{J}_j$ and confidence bounds $[L_j,U_j]$
    \State $j^+ \gets \arg\max_j \hat{J}_j$ \Comment{current leader}
    \State $j^- \gets \arg\max_{j \neq j^+} U_j$ \Comment{most competitive challenger}
    \If{$L_{j^+} \ge U_{j^-} - \varepsilon$}
        \State \Return $\combo_{j^+}$
    \EndIf
    \State Evaluate $\combo_{j^+}$ and $\combo_{j^-}$ on additional datapoints
\Until{budget exhausted}
\State \Return $\combo_{j^+}$
\end{algorithmic}
\end{algorithm}

\paragraph{Threshold Successive Elimination.}
Threshold SE is designed for settings where the user wants all combinations whose quality exceeds a target threshold $\tau$, rather than the single best combination. It classifies combinations as above threshold, below threshold, or uncertain based on confidence intervals, and removes combinations once their status is clear. 

\paragraph{Hill Climbing.}
Hill Climbing treats the search space as a discrete topology over model combinations. Starting from an initial combination, it repeatedly proposes neighboring combinations obtained by changing one role assignment at a time and moves to an improved neighbor when one is found. Multiple random restarts help escape poor local optima.

\begin{algorithm}[!ht]
\caption{Hill Climbing with random restarts}
\label{alg:hill-climb}
\begin{algorithmic}[1]
\Require Combination space $\combospace$, neighborhood function $\mathcal{N}(\cdot)$, restarts $R$
\State $\combo^\star \gets \texttt{None}$
\For{$r = 1,2,\dots,R$}
    \State Sample initial combination $\combo$
    \Repeat
        \State Evaluate neighbors $\mathcal{N}(\combo)$
        \State $\combo' \gets \arg\max_{\tilde{\combo} \in \mathcal{N}(\combo)} \hat{J}(\tilde{\combo})$
        \If{$\hat{J}(\combo') > \hat{J}(\combo)$}
            \State $\combo \gets \combo'$
        \Else
            \State stop local search
        \EndIf
    \Until{local optimum reached}
    \If{$\combo^\star = \texttt{None}$ or $\hat{J}(\combo) > \hat{J}(\combo^\star)$}
        \State $\combo^\star \gets \combo$
    \EndIf
\EndFor
\State \Return $\combo^\star$
\end{algorithmic}
\end{algorithm}

\paragraph{Bayesian Optimization.}
Bayesian Optimization~\citep{frazier2018tutorial} is useful when each end-to-end pipeline evaluation is especially expensive. It fits a surrogate model over the discrete combination space and selects the next combination via an acquisition function such as (Log) Expected Improvement~\citep{movckus1974bayesian,jones1998efficient,ament2023unexpected}. In \textsc{AgentOpt}, this corresponds to the surrogate-based search family documented alongside the other selectors.
Our implementation builds on BoTorch~\citep{balandat2020botorch}.

\begin{algorithm}[!ht]
\caption{Bayesian Optimization for model-combination search}
\label{alg:bo}
\begin{algorithmic}[1]
\Require Search space $\combospace$, initial design size $m$, total budget $B$
\State Evaluate $m$ initial combinations and collect dataset $\mathcal{H}$
\For{$t = m+1, \dots, B$}
    \State Fit surrogate model $g(\combo)$ on $\mathcal{H}$
    \State Select next combination
    \[
    \combo_t = \arg\max_{\combo \in \combospace} \alpha(\combo; g, \mathcal{H})
    \]
    where $\alpha$ is an acquisition function
    \State Evaluate $\combo_t$ and append result to $\mathcal{H}$
\EndFor
\State \Return the best observed combination in $\mathcal{H}$
\end{algorithmic}
\end{algorithm}

\paragraph{LM Proposal.}
LM Proposal uses a strong language model to propose a shortlist of promising combinations before empirical evaluation. This is appealing when semantic priors about model behavior are informative, but its reliability depends on whether those priors transfer to the role-specific interactions of the target workflow.

\subsection{Discussion}

These algorithms reflect different assumptions about the search space and evaluation process. Bandit-based methods assume that many combinations are clearly suboptimal and can be pruned quickly using partial evaluations (e.g., subsets of the data). Hill Climbing assumes that local improvements correlate with global improvement under a chosen topology and typically relies on full evaluations of each combination. Bayesian Optimization assumes enough regularity in the combination space for a surrogate model to generalize across unevaluated combinations, also based on full evaluations. Matrix UCB-E operates directly on the combination-by-datapoint evaluation grid, using upper confidence bounds to balance exploration of under-evaluated combinations with exploitation of promising ones, and can be stopped at any budget fraction to trade off cost against accuracy. In practice, this diversity is useful: it allows \textsc{AgentOpt} to support both exact search for small spaces and approximate search for larger or more expensive workflows.

\section{Experiment}
\label{sec:empirics}

\subsection{Experimental Setup}

\paragraph{Models.}
We evaluate nine models served through AWS Bedrock Application Inference Profiles using on-demand pricing from March 2026: Claude Opus 4.6 (\$5.00/\$25.00 per million input/output tokens), Claude Haiku 4.5 (\$1.00/\$5.00), Claude 3 Haiku (\$0.25/\$1.25), gpt-oss-120b (\$0.15/\$0.60), gpt-oss-20b (\$0.07/\$0.30), Kimi K2.5 (\$0.60/\$3.00), Qwen3 Next 80B A3B (\$0.15/\$1.20), Qwen3 32B (\$0.15/\$0.60), and Ministral 3 8B (\$0.15/\$0.15).

\paragraph{Benchmarks.}
We evaluate on four tasks chosen to span different workflow structures and optimization regimes. HotpotQA (199 examples) is a multi-hop question answering benchmark in the distractor setting, instantiated as a two-stage planner--solver pipeline with search tools, yielding 81 model combinations. GPQA Diamond (198 examples) is a graduate-level science multiple-choice benchmark evaluated with a single answering model, yielding 9 combinations. MathQA (200 examples) is a mathematical reasoning task instantiated as a two-stage answerer--critic pipeline with up to three retry iterations, yielding 81 combinations. BFCL v3 Multi-Turn (200 examples) evaluates multi-turn function calling with live backend state transitions and is instantiated as a single-model agent, yielding 9 combinations.

\paragraph{Agent implementation.}
All agents are implemented in LangGraph. For BFCL, models without native function-calling support, namely Qwen3 32B, Kimi K2.5, and Ministral 3 8B, are evaluated through a text-based prompting fallback. \textsc{AgentOpt} intercepts all LLM requests at the \texttt{httpx} transport layer and applies model overrides through Python \texttt{contextvars}, so no benchmark-specific agent code needs to be modified.

\paragraph{Evaluation protocol.}
We measure exact-match accuracy on HotpotQA and MathQA, multiple-choice accuracy on GPQA Diamond, and end-to-end tool-execution correctness on BFCL. For algorithmic comparisons, we report averages over 50 random seeds in order to estimate variance in search behavior.

\subsection{Results}

\paragraph{Cross-benchmark summary.}
Table~\ref{tab:summary} summarizes the best-performing combination under exhaustive search together with the cost of brute-force evaluation for each benchmark.

\begin{table}[t]
\centering
\renewcommand{\arraystretch}{1.3}
\begin{tabular}{lccccc}
\toprule
\textbf{Benchmark} & \textbf{Tuple} & \textbf{Combos} & \textbf{Best Combination} & \textbf{Acc.} & \textbf{BF Cost} \\
\midrule
GPQA Diamond & 1 & 9  & Claude Opus 4.6                                 & 74.75\% & \$4.71 \\
BFCL         & 1 & 9  & Claude Opus 4.6 (tied; Qwen3 Next: $32\times$ cheaper) & 70.00\% & \$84.80 \\
HotpotQA     & 2 & 81 & Ministral 3 8B + Claude Opus 4.6                & 74.27\% & \$51.90 \\
MathQA       & 2 & 81 & Claude Opus 4.6 + Claude Haiku 4.5            & 98.84\% & \$123.87 \\
\bottomrule
\end{tabular}
\caption{Cross-benchmark summary of exhaustive search. ``Tuple'' denotes the number of pipeline roles jointly optimized, ``Combos'' is the total number of model combinations, ``Acc.'' is the best observed task accuracy, and ``BF Cost'' is the total API cost of brute-force evaluation over the full combination space.}
\label{tab:summary}
\end{table}

Two patterns are immediately visible. First, exhaustive evaluation is already nontrivial in cost even for modest workflow sizes, especially for the two-role pipelines. Second, the best-performing combination is often not the one suggested by naive capability-based reasoning. This discrepancy is most pronounced in the multi-step settings, where role interactions determine the final outcome. Full brute force rankings for all benchmarks, including all 81 combinations for the two-role pipelines, are provided in Appendix.

\paragraph{Best combination does not equal best individual model.}
Our central empirical finding is that individual model capability does not reliably predict combination-level performance. Table~\ref{tab:surprising} highlights the most salient examples.

\begin{table}[t]
\centering
\renewcommand{\arraystretch}{1.3}
\begin{tabularx}{\textwidth}{lXX}
\toprule
\textbf{Benchmark} & \textbf{Best Combo} & \textbf{Interpretation} \\
\midrule
HotpotQA & Ministral 3 8B (planner) + Opus 4.6 (solver) & The weakest and cheapest planner performs best. When Opus is used as planner, it often bypasses the downstream solver and its search tools, leading to consistently poor end-to-end accuracy. \\
MathQA & Opus 4.6 (answer) + Haiku 4.5 (critic) & Critic identity has limited effect once the answer model is sufficiently strong. With Opus as answerer, all nine critics lie within a narrow 2.9-point range. \\
BFCL & Opus 4.6 / Kimi / Qwen3 Next (tied) & Three models achieve identical 70\% accuracy. Qwen3 Next costs 32$\times$ less than Opus, making model selection purely a cost decision at this accuracy level. \\
GPQA & Opus 4.6 (single) & This is the one setting where raw standalone capability remains predictive, and the strongest model is also the best end-to-end choice. \\
\bottomrule
\end{tabularx}
\caption{Representative cases in which the optimal model combination differs from naive capability-based expectations. The table illustrates that model quality is role-dependent and must be evaluated at the level of the full workflow rather than inferred from standalone model rankings.}
\label{tab:surprising}
\end{table}

The HotpotQA result is especially revealing. Claude Opus 4.6 is the strongest model in the benchmark by standalone capability, yet it is systematically ineffective in the planner role. All nine combinations that use Opus as planner fall in ranks 71--81 out of 81 total combinations. In seven of these nine cases, the raw execution logs carry the annotation \texttt{role2\_never\_called}, indicating that the solver is never invoked. In other words, the planner is too capable in the wrong way: it answers directly instead of delegating to the search-enabled solver. This converts model strength into a pipeline-level failure mode.

\begin{table}[t]
\centering
\renewcommand{\arraystretch}{1.2}
\begin{tabular}{rlllcc}
\toprule
\textbf{Rank} & \textbf{Planner} & \textbf{Solver} & \textbf{Acc.} & \textbf{Avg. Lat. (s)} & \textbf{Cost} \\
\midrule
71 & Claude Opus 4.6 & Kimi K2.5           & 31.96\% & 4.72 & \$2.02 \\
72 & Claude Opus 4.6 & Ministral 3 8B      & 31.96\% & 4.72 & \$2.02 \\
73 & Claude Opus 4.6 & Qwen3 32B           & 31.96\% & 4.72 & \$2.02 \\
74 & Claude Opus 4.6 & Qwen3 Next 80B A3B  & 31.96\% & 4.72 & \$2.02 \\
75 & Claude Opus 4.6 & gpt-oss-120b        & 31.95\% & 4.60 & \$2.02 \\
76 & Claude Opus 4.6 & gpt-oss-20b         & 31.88\% & 4.57 & \$2.03 \\
77 & Claude Opus 4.6 & Claude 3 Haiku      & 31.78\% & 4.22 & \$2.02 \\
78 & Claude Opus 4.6 & Claude Haiku 4.5    & 31.77\% & 4.16 & \$2.03 \\
79 & Claude Opus 4.6 & Claude Opus 4.6     & 31.71\% & 4.19 & \$2.02 \\
80 & Qwen3 32B       & Claude Haiku 4.5    & 26.63\% & 3.47 & \$0.69 \\
81 & Claude Haiku 4.5 & Claude Haiku 4.5   & 26.49\% & 3.40 & \$0.79 \\
\bottomrule
\end{tabular}
\caption{Bottom 11 combinations on HotpotQA. Most of these configurations use Claude Opus 4.6 as planner and cluster near 32\% accuracy regardless of the solver, illustrating that a strong model can be systematically mismatched to a particular pipeline role.}
\label{tab:hotpotqa-bottom}
\end{table}

Taken together, these results support the core motivation of \textsc{AgentOpt}: model quality must be evaluated in context, at the level of the full workflow, rather than inferred from standalone model rankings.

\paragraph{Algorithm comparison.}
We next compare search algorithms under a fixed evaluation budget. The full results are reported in Appendix~\ref{app:selector-results} (Tables~\ref{tab:selector-gpqa}--\ref{tab:selector-mathqa}) as 50-seed averages. Overall, Matrix UCB provides the strongest accuracy--efficiency tradeoff and is the most consistently effective method across benchmarks.

Four observations are particularly important.

\paragraph{Matrix UCB-E as the strongest selector.}
Matrix UCB-E consistently outperforms all other selectors on the larger combination spaces. At a budget fraction of 0.2, it recovers near-optimal accuracy across all four benchmarks while observing only 20\% of the evaluation grid, achieving 62--76\% cost savings relative to brute-force search. On HotpotQA it reaches 73.54\% mean accuracy (versus the 74.27\% ground truth), and on MathQA 98.37\% (versus 98.84\%), while Arm Elimination (73.19\%, 98.83\%), Hill Climbing (73.13\%, 98.76\%), and Bayesian Optimization (73.33\%, 95.41\%) all require 30--50\% more evaluations (3,356-4,635 versus 2,993-3,234) to reach comparable quality. The algorithm's advantage stems from operating directly on the combination-by-datapoint grid rather than treating each combination independently, allowing it to concentrate budget on the most promising candidates while abandoning poor ones early. 

The Low-Rank Factorization variant (Matrix UCB-E-LRF), which fits an ensemble of rank-1 approximations to predict unobserved scores, is designed to improve sample efficiency on harder problems where performance differences between combinations are subtle, by exploiting correlation structure in the scoring matrix~\citep{zhou2025speeding}. However, in our setting it consistently underperforms plain Matrix UCB-E: its rigid factorization assumption fails to capture the score patterns in larger combination spaces, and refitting the full ensemble from scratch at every step adds substantial computational overhead without improving exploration quality.

\paragraph{Sensitivity of Hill Climbing to topology.}
Hill Climbing performs well when the neighborhood structure is informative or the accuracy surface is flat. It excels on BFCL (100\% find rate, where three models tie at the top) and GPQA (where capability rankings track performance), but drops to 52\% find rate on HotpotQA, where the optimal combination is counter-intuitive and local search terminates in suboptimal regions. Arm Elimination achieves comparable mean accuracy to Hill Climbing across all four benchmarks (78.9\% versus 79.1\%) while saving modestly more evaluation budget on average (40\% versus 37\%), making it the more reliable default when the accuracy landscape is unknown in advance.

\paragraph{Failure of prior-based LM proposal.}
LM Proposal uses a strong language model to recommend combinations from model descriptions without sufficient empirical evaluation. This works on GPQA, where the best answer is intuitive and raw model capability dominates, but fails badly on HotpotQA and BFCL, where the optimal configuration depends on non-obvious role interactions. This result reinforces a central theme of the paper: in multi-step agent pipelines, reliable model selection must be empirical rather than purely inferential.
\section{Related Work}
\label{sec:related}

We review related work along four axes: agent frameworks and applications, server-side serving systems for agentic workloads, LLM routing and model selection, and bandit-based search methods for configuration optimization.

\subsection{Agent Frameworks and Applications}

The rapid maturation of LLM-based agents has been enabled by a proliferation of open-source orchestration frameworks. LangChain~\citep{mavroudis2024langchain} and its graph-based successor LangGraph~\citep{wang2024agent} provide modular abstractions for chains, tools, memory, and retrieval, and remain the most widely adopted frameworks with over 133k GitHub stars. AutoGen~\citep{wu2024autogen}, now evolving into Microsoft Agent Framework, supports multi-agent conversation patterns and flexible role-based interaction. CrewAI~\citep{venkadesh2024unlocking} emphasizes role-based collaboration that mirrors human team structures, while MetaGPT~\citep{hong2023metagpt} uses standardized operating procedures to coordinate specialized agents in software engineering workflows. On the commercial side, systems like Manus and ClaudeCode~\citep{chatlatanagulchai2025use} demonstrate that agentic capabilities have moved well beyond research prototypes into production use.

These frameworks share a common architectural pattern: an LLM serves as the cognitive controller that orchestrates planning, tool use, and multi-step reasoning over extended trajectories. Recent surveys~\citep{xi2023rise} provide broad taxonomies of agent architectures, decomposing them into perception, memory, planning, and action modules. More recent work~\citep{masterman2024landscape} further classifies agents by interaction topology, such as chain, star, mesh, and explicit workflow graphs, highlighting the growing diversity of multi-agent designs. The transition from single-agent loops to multi-agent systems has been dramatic: Gartner reported a 1,445\% surge in multi-agent system inquiries from Q1 2024 to Q2 2025.

A key insight from this literature is that agent performance depends critically on the interplay between the orchestration logic and the underlying model capabilities. \citet{belcak2025small} argue that small language models can replace LLMs in roughly 60\% of agent queries in frameworks like MetaGPT, suggesting that heterogeneous model assignment across agent roles is both natural and economically motivated. Our work builds directly on this observation: rather than treating model selection as a fixed design choice, \textsc{AgentOpt} treats it as an optimization problem over the combinatorial space of role-to-model assignments.

\subsection{Server-Side Agent Serving Systems}

As agentic workloads have scaled, a new class of serving systems has emerged to optimize their execution on the provider side. These systems extend traditional LLM serving engines such as vLLM~\citep{kwon2023vllm} and SGLang~\citep{zheng2024sglang}, which optimize individual inference requests, to reason about the multi-turn, tool-interleaved structure of agent programs.

\textbf{Autellix}~\citep{luo2025autellix} was among the first to treat agentic programs as first-class scheduling citizens. It intercepts LLM calls submitted by agent programs and enriches schedulers with program-level context, proposing scheduling algorithms for both single-threaded and distributed programs that preempt and prioritize calls based on cumulative service time. Autellix demonstrates 4--15$\times$ throughput improvements over vanilla vLLM by reducing head-of-line blocking at both the request and program levels.

\textbf{Continuum}~\citep{li2025continuum} addresses a different bottleneck: KV cache management during tool-call pauses. In multi-turn agent workflows, each tool invocation creates a pause that can trigger KV cache eviction, forcing expensive recomputation on subsequent turns. Continuum introduces a KV cache time-to-live (TTL) mechanism that predicts tool-call durations and selectively pins KV caches in GPU memory, combined with program-level first-come-first-serve scheduling to prevent scheduling bubbles.

\textbf{ThunderAgent}~\citep{kang2026thunderagent} unifies the treatment of heterogeneous resources, including KV caches, system states, and external tool assets such as Docker containers and network ports, under a single program abstraction. Its program-aware scheduler maximizes KV cache hit rates while a tool resource manager handles lifecycle management, preventing resource leaks from zombie tool environments. ThunderAgent reports 1.5--3.6$\times$ throughput improvements for agentic serving and up to 4.2$\times$ disk memory savings.

\textbf{AIOS}~\citep{mei2024aios} takes a broader systems perspective, proposing an operating-system-like architecture that isolates LLM-specific services (scheduling, context management, memory management, access control) into a dedicated kernel layer. AIOS supports agents built from diverse frameworks including ReAct~\citep{yao2022react}, Reflexion~\citep{shinn2023reflexion}, AutoGen, and MetaGPT~\citep{hong2023metagpt}, providing a unified runtime for concurrent agent execution.

Other systems address complementary aspects of the agentic serving stack. InferCept~\citep{abhyankar2024infercept} introduces selective KV cache preservation during tool calls. Parrot~\citep{lin2024parrot} and Alto~\citep{santhanam2024alto} optimize for static workflow structures. Tempo~\citep{zhang2025tempo} proposes SLO-aware scheduling that differentiates between chat, agent, and reasoning request types.

All of these systems optimize the \emph{execution} of a fixed agent pipeline: given a model assignment, they minimize latency, maximize throughput, or improve resource utilization on the server side. \textsc{AgentOpt} operates on a fundamentally different axis. Rather than optimizing how a given model assignment is served, it optimizes \emph{which} model assignment to use in the first place. The two approaches are complementary: server-side serving improvements apply regardless of model choice, while client-side model selection determines the quality-cost-latency frontier that serving systems operate within.

\subsection{LLM Routing and Model Selection}

LLM routing assigns incoming queries to the most suitable model from a candidate pool, balancing performance against cost and latency. RouterBench~\citep{hu2024routerbench} established standardized evaluation for routing systems, providing over 405K inference outcomes across representative LLMs. RouterEval~\citep{huang2025routereval} scaled this effort to over 8,500 LLMs and 200 million performance records, revealing a model-level scaling phenomenon where capable routers can surpass the performance of any individual model in the pool.

Routing methods span several paradigms. Semantic and intent-based routers use query embeddings to predict model suitability: RouteLLM~\citep{ong2024routellm} leverages human preference data, while others employ retrieval-based patterns or lightweight encoders such as DeBERTa. More recent work explores mechanistic routing using internal hidden states, showing that LLMs encode accuracy predictions within the residual stream during prefill. OmniRouter~\citep{mei2025omnirouter} and hybrid approaches combine multiple routing signals for improved selection.

Router-R1~\citep{zhang2025router} moves beyond single-round, one-to-one routing by formulating multi-LLM routing as a sequential decision process. Using reinforcement learning, it interleaves reasoning with dynamic model invocation and aggregates responses across models, representing an important step toward multi-step routing. However, Router-R1 still operates at the level of individual queries rather than optimizing over full pipeline trajectories.

The critical distinction between existing LLM routing and our work is the \emph{unit of optimization}. Standard routing methods make per-query, per-call decisions: given a single input, select the best model. In multi-step agent pipelines, however, model assignments are coupled across stages. A planner's output conditions the solver's effectiveness, and a solver's behavior depends on the tools and context established by upstream agents. As we show empirically, the optimal model for one role depends on which models occupy other roles, making per-call routing fundamentally insufficient. \textsc{AgentOpt} addresses this by optimizing over full model \emph{combinations}, treating each assignment of models to pipeline roles as an atomic unit of evaluation.

\subsection{Bandit Methods and Configuration Search}

The problem of selecting the best model combination from a combinatorial space connects to a rich literature on multi-armed bandits and automated machine learning. Hyperband~\citep{li2018hyperband} formulates hyperparameter optimization as a pure-exploration bandit problem, using adaptive resource allocation and early stopping to achieve order-of-magnitude speedups over Bayesian optimization. Successive Halving~\citep{jamieson2016non} provides the algorithmic foundation, progressively eliminating poorly performing configurations by allocating increasing budgets to surviving candidates. Recent work has also applied bandit algorithms (e.g., UCB variants) to model selection for large language models~\citep{zhou2024speeding}, though these approaches focus on selecting among individual models rather than optimizing full agent workflows.

In the AutoML literature, the Combined Algorithm Selection and Hyperparameter optimization (CASH) problem~\citep{guo2019new} jointly searches over model classes and their hyperparameters. 
Our setting differs from classical CASH in two important ways. First, each ``arm'' in our formulation corresponds to a full model combination rather than a single algorithm with hyperparameters. The reward of each arm is the end-to-end pipeline performance, which can only be observed after executing the complete agent trajectory, not incrementally as in iterative training. Second, our objective is multi-dimensional: we seek the Pareto frontier over accuracy, cost, and latency, rather than optimizing a single scalar metric. These differences motivate the use of arm elimination methods~\citep{wen2025optimal, shahrampour2017sequential, qian2016randomized} and matrix-based UCB exploration in our setting. In particular, treating the combination-by-datapoint evaluations as a partially observed matrix allows UCB-based selection to exploit the grid structure directly, concentrating budget on promising combinations while observing only a fraction of the full evaluation space, leading to substantial empirical improvements in evaluation efficiency.
\section{Conclusion}
\label{sec:conclusion}

As AI agents become increasingly customizable and widely deployed, efficiency optimization can no longer be treated solely as a server-side systems problem. Developers now construct agentic workflows from heterogeneous models, tools, and APIs, and many of the most important optimization decisions are made on the client side. These decisions are inherently application-dependent, reflecting task-specific preferences over performance, latency, and monetary cost. This makes client-side optimization a distinct and necessary layer of the agent systems stack.

In this report, we introduced \textsc{AgentOpt}, a framework-agnostic Python package for client-side optimization of agentic workflows, with model selection as its first focus. We argued that model selection is a first-class optimization problem because it determines the computational substrate on which all other optimizations operate, and because the empirical differences between model combinations can be dramatic. Across our benchmarks, the cost gap between near-equivalent combinations ranges from 13$\times$ to 32$\times$, while the best end-to-end configuration often differs sharply from what standalone model rankings would suggest. These results show that model quality must be evaluated in context, at the level of the full workflow.

We further framed agent model selection as an end-to-end sequential decision problem, naturally connected to MDP-style reasoning, and showed that practical optimization requires efficient search over a combinatorial space of model assignments. To make this feasible, \textsc{AgentOpt} combines a lightweight, framework-agnostic execution substrate with multiple search strategies, including elimination-based methods, local search, and surrogate-based optimization. Among these, Arm Elimination provides the most consistent tradeoff between search cost and solution quality across benchmarks.

More broadly, the package is intended as a foundation for a wider class of client-side optimizations beyond static model assignment, including adaptive routing, tool selection, scheduling, and personalized utility-aware policies. We view this as the next stage of agent systems research: not only serving agents efficiently at scale, but giving developers direct and principled control over how their own agents trade off quality, latency, and cost in deployment.

The framework is available as an open-source Python library at \url{https://github.com/AgentOptimizer/agentopt}.

\bibliographystyle{plainnat}
\bibliography{references}

\appendix
\section{Full Benchmark Results}
\label{app:full-results}

\setlength{\floatsep}{8pt}
\setlength{\textfloatsep}{10pt}
\setlength{\intextsep}{8pt}

\subsection{Selector Comparison by Benchmark}
\label{app:selector-results}

All selector results averaged over 50 random seeds.

\begin{table}[!htbp]
\centering
\caption{GPQA Diamond selector comparison (198 samples, 9 models).}
\label{tab:selector-gpqa}
\renewcommand{\arraystretch}{1.2}
\begin{tabular}{lccccc}
\toprule
\textbf{Selector} & \textbf{Budget} & \textbf{Mean Acc} & \textbf{Mean Evals} & \textbf{Mean Cost} & \textbf{Cost Savings} \\
\midrule
Brute Force (ref) & -- & 74.75\% & 1,782 & \$4.71 & 0.0\% \\
Matrix UCB-E & 0.5 & 74.75\% & 891 & \$4.18 & 11.4\% \\
Matrix UCB-E-LRF & 0.5 & 74.75\% & 891 & \$4.17 & 11.6\% \\
LM Proposal & -- & 74.75\% & 198 & \$2.47 & 47.6\% \\
Hill Climbing & -- & 74.55\% & 1,501 & \$4.03 & 14.4\% \\
Matrix UCB-E & 0.3 & 74.34\% & 535 & \$3.11 & 33.9\% \\
Arm Elimination & -- & 74.10\% & 666 & \$3.57 & 24.3\% \\
Matrix UCB-E & 0.2 & 73.87\% & 357 & \$1.79 & 61.9\% \\
Matrix UCB-E-LRF & 0.2 & 73.70\% & 357 & \$1.12 & 76.2\% \\
Epsilon-LUCB & -- & 73.14\% & 380 & \$2.51 & 46.7\% \\
Matrix UCB-E-LRF & 0.3 & 72.62\% & 535 & \$2.57 & 45.5\% \\
Bayesian Opt & -- & 72.43\% & 990 & \$2.59 & 45.0\% \\
Matrix UCB-E-LRF & 0.1 & 71.43\% & 179 & \$0.48 & 89.8\% \\
Matrix UCB-E & 0.1 & 71.28\% & 179 & \$0.49 & 89.5\% \\
Random Search & -- & 68.57\% & 594 & \$1.73 & 63.3\% \\
Threshold SE & -- & 57.48\% & 252 & \$1.80 & 61.8\% \\
\bottomrule
\end{tabular}
\end{table}
\begin{table}[!htbp]
\centering
\caption{BFCL v3 selector comparison (200 samples, 9 models).}
\label{tab:selector-bfcl}
\renewcommand{\arraystretch}{1.2}
\begin{tabular}{lccccc}
\toprule
\textbf{Selector} & \textbf{Budget} & \textbf{Mean Acc} & \textbf{Mean Evals} & \textbf{Mean Cost} & \textbf{Cost Savings} \\
\midrule
Brute Force (ref) & -- & 70.00\% & 1,800 & \$84.80 & 0.0\% \\
Hill Climbing & -- & 70.00\% & 1,664 & \$72.12 & 15.0\% \\
Matrix UCB-E & 0.5 & 70.00\% & 900 & \$73.10 & 13.8\% \\
Matrix UCB-E & 0.2 & 69.90\% & 360 & \$26.14 & 69.2\% \\
Matrix UCB-E & 0.3 & 69.90\% & 540 & \$43.43 & 48.8\% \\
Epsilon-LUCB & -- & 69.90\% & 399 & \$40.03 & 52.8\% \\
Arm Elimination & -- & 69.37\% & 912 & \$74.39 & 12.3\% \\
Matrix UCB-E-LRF & 0.2 & 69.30\% & 360 & \$19.77 & 76.7\% \\
Matrix UCB-E-LRF & 0.3 & 69.30\% & 540 & \$38.64 & 54.4\% \\
Bayesian Opt & -- & 69.27\% & 1,000 & \$50.64 & 40.3\% \\
Matrix UCB-E & 0.1 & 68.87\% & 180 & \$8.46 & 90.0\% \\
Matrix UCB-E-LRF & 0.1 & 68.15\% & 180 & \$8.98 & 89.4\% \\
Random Search & -- & 67.13\% & 600 & \$31.39 & 63.0\% \\
Matrix UCB-E-LRF & 0.5 & 67.04\% & 900 & \$70.90 & 16.4\% \\
Threshold SE & -- & 58.19\% & 186 & \$18.82 & 77.8\% \\
LM Proposal & -- & 44.03\% & 200 & \$3.39 & 96.0\% \\
\bottomrule
\end{tabular}
\end{table}

\clearpage

\begin{table}[!htbp]
\centering
\caption{HotpotQA selector comparison (199 samples, 81 combinations).}
\label{tab:selector-hotpotqa}
\renewcommand{\arraystretch}{1.2}
\begin{tabular}{lccccc}
\toprule
\textbf{Selector} & \textbf{Budget} & \textbf{Mean Acc} & \textbf{Mean Evals} & \textbf{Mean Cost} & \textbf{Cost Savings} \\
\midrule
Brute Force (ref) & -- & 74.27\% & 16,168 & \$51.90 & 0.0\% \\
Matrix UCB-E & 0.5 & 74.27\% & 8,084 & \$23.18 & 55.3\% \\
Matrix UCB-E & 0.3 & 74.25\% & 4,851 & \$18.41 & 64.5\% \\
Matrix UCB-E & 0.2 & 73.54\% & 3,234 & \$12.49 & 75.9\% \\
Bayesian Opt & -- & 73.33\% & 3,996 & \$12.29 & 76.3\% \\
Arm Elimination & -- & 73.19\% & 4,283 & \$16.92 & 67.4\% \\
Hill Climbing & -- & 73.13\% & 4,635 & \$19.39 & 62.6\% \\
Random Search & -- & 72.25\% & 4,192 & \$13.37 & 74.2\% \\
Epsilon-LUCB & -- & 69.71\% & 478 & \$1.75 & 96.6\% \\
Matrix UCB-E & 0.1 & 69.54\% & 1,617 & \$5.18 & 90.0\% \\
Matrix UCB-E-LRF & 0.1 & 67.83\% & 1,617 & \$5.97 & 88.5\% \\
Threshold SE & -- & 65.42\% & 1,642 & \$6.45 & 87.6\% \\
Matrix UCB-E-LRF & 0.2 & 65.29\% & 3,234 & \$14.96 & 71.2\% \\
Matrix UCB-E-LRF & 0.5 & 61.51\% & 8,084 & \$23.11 & 55.5\% \\
Matrix UCB-E-LRF & 0.3 & 59.81\% & 4,851 & \$18.85 & 63.7\% \\
LM Proposal & -- & 34.13\% & 200 & \$1.84 & 96.5\% \\
\bottomrule
\end{tabular}
\end{table}

\begin{table}[!htbp]
\centering
\caption{MathQA selector comparison (200 samples, 81 combinations).}
\label{tab:selector-mathqa}
\renewcommand{\arraystretch}{1.2}
\begin{tabular}{lccccc}
\toprule
\textbf{Selector} & \textbf{Budget} & \textbf{Mean Acc} & \textbf{Mean Evals} & \textbf{Mean Cost} & \textbf{Cost Savings} \\
\midrule
Brute Force (ref) & -- & 98.84\% & 14,961 & \$123.87 & 0.0\% \\
Matrix UCB-E & 0.5 & 98.84\% & 7,481 & \$89.10 & 28.1\% \\
Matrix UCB-E-LRF & 0.5 & 98.84\% & 7,481 & \$89.48 & 27.8\% \\
Matrix UCB-E & 0.3 & 98.84\% & 4,489 & \$62.12 & 49.9\% \\
Arm Elimination & -- & 98.83\% & 3,356 & \$51.86 & 58.1\% \\
Hill Climbing & -- & 98.76\% & 3,926 & \$54.22 & 56.2\% \\
Matrix UCB-E & 0.2 & 98.37\% & 2,993 & \$35.18 & 71.6\% \\
Random Search & -- & 98.17\% & 3,880 & \$31.77 & 74.4\% \\
Epsilon-LUCB & -- & 96.99\% & 447 & \$6.10 & 95.1\% \\
LM Proposal & -- & 95.82\% & 158 & \$5.61 & 95.5\% \\
Matrix UCB-E & 0.1 & 95.57\% & 1,497 & \$13.50 & 89.1\% \\
Bayesian Opt & -- & 95.41\% & 3,666 & \$35.56 & 71.3\% \\
Matrix UCB-E-LRF & 0.3 & 93.57\% & 4,489 & \$65.98 & 46.7\% \\
Matrix UCB-E-LRF & 0.2 & 91.90\% & 2,993 & \$40.64 & 67.2\% \\
Matrix UCB-E-LRF & 0.1 & 90.28\% & 1,497 & \$16.85 & 86.4\% \\
Threshold SE & -- & 74.52\% & 1,355 & \$6.90 & 94.4\% \\
\bottomrule
\end{tabular}
\end{table}
\clearpage

\subsection{Benchmark Brute Force Results}
\label{app:brute-force}
\begin{table}[H]
\centering
\caption{GPQA Diamond brute force results (198 samples, 9 models).}
\label{tab:bf-gpqa}
\renewcommand{\arraystretch}{1.1}
\begin{tabular}{clccc}
\toprule
\textbf{Rank} & \textbf{Model} & \textbf{Accuracy} & \textbf{Avg. Lat. (s)} & \textbf{Cost} \\
\midrule
1 & Claude Opus 4.6 & 74.75\% & 9.16 & \$2.47 \\
2 & Kimi K2.5 & 72.73\% & 16.41 & \$1.13 \\
3 & gpt-oss-120b & 68.18\% & 6.46 & \$0.19 \\
4 & Claude Haiku 4.5 & 59.60\% & 3.70 & \$0.52 \\
5 & Qwen3 Next 80B A3B & 51.01\% & 10.33 & \$0.13 \\
6 & gpt-oss-20b & 50.00\% & 6.21 & \$0.13 \\
7 & Qwen3 32B & 46.97\% & 1.54 & \$0.07 \\
8 & Ministral 3 8B & 36.87\% & 0.25 & \$0.007 \\
9 & Claude 3 Haiku & 34.85\% & 1.79 & \$0.056 \\
\bottomrule
\end{tabular}

\vspace{0.5em}

\caption{BFCL v3 brute force results (200 samples, 9 models).}
\label{tab:bf-bfcl}
\renewcommand{\arraystretch}{1.1}
\begin{tabular}{clcccc}
\toprule
\textbf{Rank} & \textbf{Model} & \textbf{Accuracy} & \textbf{Avg. Lat. (s)} & \textbf{Avg Calls} & \textbf{Cost} \\
\midrule
1 & Claude Opus 4.6 & 70.00\% & 42.35 & 12.2 & \$60.13 \\
2 & Kimi K2.5 & 70.00\% & 21.30 & 12.0 & \$3.85 \\
3 & Qwen3 Next 80B A3B & 70.00\% & 60.54 & 15.3 & \$1.90 \\
4 & Claude Haiku 4.5 & 65.00\% & 20.90 & 11.1 & \$11.98 \\
5 & gpt-oss-120b & 58.50\% & 20.01 & 14.5 & \$1.16 \\
6 & Qwen3 32B & 47.00\% & 10.78 & 11.9 & \$1.01 \\
7 & Claude 3 Haiku & 43.50\% & 17.96 & 16.4 & \$3.42 \\
8 & gpt-oss-20b & 42.00\% & 10.03 & 9.8 & \$0.43 \\
9 & Ministral 3 8B & 34.00\% & 29.03 & 9.7 & \$0.93 \\
\bottomrule
\end{tabular}
\end{table}

\clearpage
\thispagestyle{empty}
\begin{table}[p]
\centering
\caption{HotpotQA brute force results (199 samples, 81 planner--solver combinations).}
\label{tab:bf-hotpotqa}
\renewcommand{\arraystretch}{1.05}
\scriptsize
\begin{tabular}{cllccc}
\toprule
\textbf{Rank} & \textbf{Planner} & \textbf{Solver} & \textbf{Acc} & \textbf{Avg. Lat. (s)} & \textbf{Cost} \\
\midrule
1 & Ministral 3 8B & Claude Opus 4.6 & 74.27\% & 4.97 & \$2.64 \\
2 & Claude 3 Haiku & Claude Opus 4.6 & 73.25\% & 4.52 & \$2.79 \\
3 & Qwen3 32B & Claude Opus 4.6 & 73.02\% & 4.26 & \$2.65 \\
4 & Qwen3 Next 80B A3B & Claude Opus 4.6 & 72.10\% & 4.67 & \$2.67 \\
5 & Qwen3 Next 80B A3B & gpt-oss-120b & 71.83\% & 3.07 & \$0.13 \\
6 & Qwen3 32B & gpt-oss-120b & 70.04\% & 2.66 & \$0.13 \\
7 & Kimi K2.5 & Claude Opus 4.6 & 69.96\% & 4.49 & \$2.43 \\
8 & Claude 3 Haiku & gpt-oss-120b & 69.86\% & 3.21 & \$0.17 \\
9 & Ministral 3 8B & gpt-oss-20b & 69.34\% & 5.66 & \$0.09 \\
10 & Claude 3 Haiku & Qwen3 Next 80B A3B & 69.27\% & 3.00 & \$0.16 \\
11 & Qwen3 Next 80B A3B & gpt-oss-20b & 68.89\% & 2.82 & \$0.09 \\
12 & Ministral 3 8B & gpt-oss-120b & 68.70\% & 3.65 & \$0.12 \\
13 & Qwen3 Next 80B A3B & Qwen3 Next 80B A3B & 68.15\% & 2.69 & \$0.11 \\
14 & Ministral 3 8B & Qwen3 Next 80B A3B & 67.98\% & 3.85 & \$0.11 \\
15 & Qwen3 32B & Qwen3 Next 80B A3B & 67.53\% & 3.51 & \$0.11 \\
16 & Qwen3 32B & gpt-oss-20b & 66.95\% & 2.48 & \$0.09 \\
17 & Claude 3 Haiku & Ministral 3 8B & 65.98\% & 3.73 & \$0.14 \\
18 & Ministral 3 8B & Kimi K2.5 & 65.24\% & 3.27 & \$0.26 \\
19 & gpt-oss-120b & Qwen3 Next 80B A3B & 64.93\% & 4.68 & \$0.10 \\
20 & Ministral 3 8B & Ministral 3 8B & 64.89\% & 3.55 & \$0.09 \\
21 & Claude 3 Haiku & gpt-oss-20b & 64.79\% & 2.90 & \$0.13 \\
22 & Kimi K2.5 & gpt-oss-120b & 64.70\% & 4.16 & \$0.29 \\
23 & gpt-oss-120b & Claude Opus 4.6 & 64.59\% & 4.57 & \$1.61 \\
24 & gpt-oss-120b & Claude Haiku 4.5 & 64.11\% & 4.26 & \$0.38 \\
25 & Kimi K2.5 & Qwen3 Next 80B A3B & 63.99\% & 4.39 & \$0.30 \\
26 & Kimi K2.5 & Ministral 3 8B & 63.95\% & 6.42 & \$0.28 \\
27 & Claude 3 Haiku & Kimi K2.5 & 63.85\% & 2.89 & \$0.31 \\
28 & gpt-oss-120b & Ministral 3 8B & 63.70\% & 7.37 & \$0.09 \\
29 & Qwen3 Next 80B A3B & Kimi K2.5 & 63.69\% & 2.89 & \$0.27 \\
30 & Kimi K2.5 & gpt-oss-20b & 63.35\% & 6.80 & \$0.26 \\
31 & Qwen3 32B & Kimi K2.5 & 63.17\% & 3.26 & \$0.28 \\
32 & gpt-oss-120b & Claude 3 Haiku & 62.72\% & 3.72 & \$0.13 \\
33 & Kimi K2.5 & Kimi K2.5 & 62.28\% & 4.56 & \$0.44 \\
34 & gpt-oss-120b & gpt-oss-120b & 62.15\% & 4.59 & \$0.10 \\
35 & Qwen3 Next 80B A3B & Ministral 3 8B & 62.11\% & 4.27 & \$0.10 \\
36 & gpt-oss-120b & gpt-oss-20b & 61.51\% & 2.71 & \$0.08 \\
37 & Qwen3 32B & Ministral 3 8B & 61.17\% & 2.89 & \$0.09 \\
38 & gpt-oss-120b & Kimi K2.5 & 60.85\% & 4.09 & \$0.18 \\
39 & gpt-oss-120b & Qwen3 32B & 58.80\% & 4.06 & \$0.10 \\
40 & Claude 3 Haiku & Qwen3 32B & 56.02\% & 2.87 & \$0.15 \\
41 & Claude 3 Haiku & Claude 3 Haiku & 55.91\% & 2.41 & \$0.21 \\
42 & gpt-oss-20b & Claude Opus 4.6 & 55.86\% & 2.84 & \$1.04 \\
43 & Ministral 3 8B & Qwen3 32B & 55.02\% & 3.63 & \$0.11 \\
44 & Kimi K2.5 & Claude 3 Haiku & 54.90\% & 3.42 & \$0.34 \\
45 & Qwen3 32B & Qwen3 32B & 54.82\% & 2.53 & \$0.11 \\
46 & Kimi K2.5 & Qwen3 32B & 54.73\% & 4.57 & \$0.30 \\
47 & gpt-oss-20b & Claude Haiku 4.5 & 54.28\% & 2.19 & \$0.26 \\
48 & gpt-oss-20b & Ministral 3 8B & 54.25\% & 4.35 & \$0.05 \\
49 & Qwen3 Next 80B A3B & Qwen3 32B & 54.13\% & 2.83 & \$0.11 \\
50 & gpt-oss-20b & Qwen3 Next 80B A3B & 53.89\% & 2.11 & \$0.06 \\
51 & gpt-oss-20b & Claude 3 Haiku & 52.66\% & 2.04 & \$0.08 \\
52 & gpt-oss-20b & gpt-oss-120b & 52.17\% & 2.11 & \$0.06 \\
53 & Ministral 3 8B & Claude 3 Haiku & 51.33\% & 4.10 & \$0.16 \\
54 & gpt-oss-20b & Kimi K2.5 & 51.01\% & 1.96 & \$0.12 \\
55 & gpt-oss-20b & gpt-oss-20b & 50.09\% & 2.12 & \$0.05 \\
56 & Qwen3 Next 80B A3B & Claude 3 Haiku & 49.98\% & 2.56 & \$0.17 \\
57 & gpt-oss-20b & Qwen3 32B & 49.16\% & 2.05 & \$0.06 \\
58 & Qwen3 32B & Claude 3 Haiku & 48.77\% & 2.23 & \$0.16 \\
59 & Claude 3 Haiku & Claude Haiku 4.5 & 46.50\% & 3.35 & \$0.71 \\
60 & Claude Haiku 4.5 & Claude Opus 4.6 & 43.54\% & 4.06 & \$1.80 \\
61 & Claude Haiku 4.5 & gpt-oss-20b & 41.49\% & 3.03 & \$0.45 \\
62 & Claude Haiku 4.5 & gpt-oss-120b & 41.20\% & 3.14 & \$0.47 \\
63 & Claude Haiku 4.5 & Qwen3 Next 80B A3B & 41.17\% & 2.95 & \$0.46 \\
64 & Claude Haiku 4.5 & Ministral 3 8B & 41.09\% & 3.75 & \$0.45 \\
65 & Claude Haiku 4.5 & Kimi K2.5 & 41.00\% & 6.16 & \$0.54 \\
66 & Kimi K2.5 & Claude Haiku 4.5 & 37.19\% & 4.23 & \$0.88 \\
67 & Claude Haiku 4.5 & Qwen3 32B & 36.13\% & 2.89 & \$0.46 \\
68 & Claude Haiku 4.5 & Claude 3 Haiku & 34.34\% & 2.63 & \$0.49 \\
69 & Ministral 3 8B & Claude Haiku 4.5 & 32.42\% & 4.14 & \$0.70 \\
70 & Qwen3 Next 80B A3B & Claude Haiku 4.5 & 32.19\% & 3.92 & \$0.72 \\
71 & Claude Opus 4.6 & Kimi K2.5 & 31.96\% & 4.72 & \$2.02 \\
72 & Claude Opus 4.6 & Ministral 3 8B & 31.96\% & 4.72 & \$2.02 \\
73 & Claude Opus 4.6 & Qwen3 32B & 31.96\% & 4.72 & \$2.02 \\
74 & Claude Opus 4.6 & Qwen3 Next 80B A3B & 31.96\% & 4.72 & \$2.02 \\
75 & Claude Opus 4.6 & gpt-oss-120b & 31.95\% & 4.60 & \$2.02 \\
76 & Claude Opus 4.6 & gpt-oss-20b & 31.88\% & 4.57 & \$2.03 \\
77 & Claude Opus 4.6 & Claude 3 Haiku & 31.78\% & 4.22 & \$2.02 \\
78 & Claude Opus 4.6 & Claude Haiku 4.5 & 31.77\% & 4.16 & \$2.03 \\
79 & Claude Opus 4.6 & Claude Opus 4.6 & 31.71\% & 4.19 & \$2.02 \\
80 & Qwen3 32B & Claude Haiku 4.5 & 26.63\% & 3.47 & \$0.69 \\
81 & Claude Haiku 4.5 & Claude Haiku 4.5 & 26.49\% & 3.40 & \$0.79 \\
\bottomrule
\end{tabular}
\end{table}


\clearpage
\thispagestyle{empty}
\begin{table}[p]
\centering
\caption{MathQA brute force results (200 samples, 81 answer--critic combinations).}
\label{tab:bf-mathqa}
\renewcommand{\arraystretch}{1.05}
\scriptsize
\begin{tabular}{cllccc}
\toprule
\textbf{Rank} & \textbf{Answer Model} & \textbf{Critic Model} & \textbf{Acc} & \textbf{Avg. Lat. (s)} & \textbf{Cost} \\
\midrule
1 & Claude Opus 4.6 & Claude Haiku 4.5 & 98.84\% & 16.15 & \$6.19 \\
2 & Claude Opus 4.6 & Qwen3 Next 80B A3B & 98.82\% & 14.30 & \$5.77 \\
3 & Claude Opus 4.6 & Ministral 3 8B & 98.72\% & 14.03 & \$5.26 \\
4 & Claude Opus 4.6 & gpt-oss-20b & 98.28\% & 16.50 & \$5.93 \\
5 & Claude Opus 4.6 & gpt-oss-120b & 97.77\% & 15.40 & \$6.30 \\
6 & Claude Opus 4.6 & Qwen3 32B & 97.28\% & 15.05 & \$6.68 \\
7 & Claude Opus 4.6 & Claude Opus 4.6 & 97.24\% & 15.94 & \$6.97 \\
8 & Claude Opus 4.6 & Kimi K2.5 & 97.24\% & 18.37 & \$6.58 \\
9 & Claude Opus 4.6 & Claude 3 Haiku & 95.95\% & 14.85 & \$5.37 \\
10 & gpt-oss-20b & Claude Opus 4.6 & 94.57\% & 6.81 & \$0.97 \\
11 & gpt-oss-20b & Kimi K2.5 & 94.57\% & 12.45 & \$0.26 \\
12 & gpt-oss-20b & gpt-oss-20b & 94.54\% & 4.04 & \$0.08 \\
13 & Claude Haiku 4.5 & Qwen3 32B & 94.50\% & 12.68 & \$2.51 \\
14 & gpt-oss-20b & Claude Haiku 4.5 & 94.05\% & 6.19 & \$0.37 \\
15 & gpt-oss-20b & gpt-oss-120b & 94.02\% & 4.94 & \$0.11 \\
16 & gpt-oss-20b & Qwen3 Next 80B A3B & 94.02\% & 8.67 & \$0.14 \\
17 & Claude Haiku 4.5 & Claude Haiku 4.5 & 94.00\% & 14.31 & \$2.59 \\
18 & gpt-oss-20b & Ministral 3 8B & 93.99\% & 8.27 & \$0.10 \\
19 & gpt-oss-120b & Claude Opus 4.6 & 93.81\% & 9.10 & \$1.25 \\
20 & Claude Haiku 4.5 & gpt-oss-20b & 93.50\% & 12.51 & \$2.20 \\
21 & Claude Haiku 4.5 & Claude Opus 4.6 & 93.50\% & 15.82 & \$3.77 \\
22 & Claude Haiku 4.5 & Ministral 3 8B & 93.50\% & 14.70 & \$2.57 \\
23 & Claude Haiku 4.5 & Kimi K2.5 & 93.50\% & 17.50 & \$2.60 \\
24 & gpt-oss-20b & Qwen3 32B & 93.48\% & 4.30 & \$0.09 \\
25 & gpt-oss-20b & Claude 3 Haiku & 93.44\% & 6.10 & \$0.15 \\
26 & gpt-oss-120b & Ministral 3 8B & 93.26\% & 10.42 & \$0.19 \\
27 & gpt-oss-120b & Qwen3 32B & 93.26\% & 5.53 & \$0.16 \\
28 & Claude Haiku 4.5 & gpt-oss-120b & 93.00\% & 14.65 & \$2.90 \\
29 & Claude Haiku 4.5 & Qwen3 Next 80B A3B & 93.00\% & 20.98 & \$7.81 \\
30 & gpt-oss-120b & Claude Haiku 4.5 & 92.82\% & 7.77 & \$0.47 \\
31 & gpt-oss-120b & gpt-oss-20b & 92.78\% & 6.45 & \$0.18 \\
32 & gpt-oss-120b & gpt-oss-120b & 92.78\% & 6.94 & \$0.19 \\
33 & gpt-oss-120b & Kimi K2.5 & 92.78\% & 12.09 & \$0.32 \\
34 & gpt-oss-120b & Qwen3 Next 80B A3B & 92.78\% & 10.98 & \$0.23 \\
35 & gpt-oss-120b & Claude 3 Haiku & 92.75\% & 6.42 & \$0.20 \\
36 & Claude Haiku 4.5 & Claude 3 Haiku & 92.50\% & 13.43 & \$2.46 \\
37 & Claude 3 Haiku & Claude Opus 4.6 & 89.66\% & 13.32 & \$2.26 \\
38 & Qwen3 32B & Qwen3 Next 80B A3B & 88.83\% & 8.02 & \$0.24 \\
39 & Ministral 3 8B & Claude 3 Haiku & 88.15\% & 10.24 & \$0.05 \\
40 & Qwen3 32B & gpt-oss-120b & 87.83\% & 7.11 & \$0.47 \\
41 & Ministral 3 8B & Qwen3 Next 80B A3B & 87.82\% & 9.22 & \$0.03 \\
42 & Qwen3 32B & Claude Opus 4.6 & 87.56\% & 12.33 & \$3.43 \\
43 & Ministral 3 8B & Kimi K2.5 & 87.04\% & 14.43 & \$0.09 \\
44 & Ministral 3 8B & gpt-oss-120b & 86.63\% & 10.58 & \$0.07 \\
45 & Claude 3 Haiku & Claude Haiku 4.5 & 86.55\% & 9.32 & \$0.69 \\
46 & Ministral 3 8B & Ministral 3 8B & 86.52\% & 7.29 & \$0.03 \\
47 & Ministral 3 8B & Claude Opus 4.6 & 86.47\% & 11.46 & \$0.93 \\
48 & Qwen3 32B & Claude Haiku 4.5 & 86.46\% & 7.47 & \$0.90 \\
49 & Ministral 3 8B & Claude Haiku 4.5 & 86.23\% & 11.66 & \$0.30 \\
50 & Ministral 3 8B & gpt-oss-20b & 86.13\% & 12.33 & \$0.05 \\
51 & Qwen3 32B & Ministral 3 8B & 86.10\% & 17.57 & \$0.21 \\
52 & Qwen3 32B & Kimi K2.5 & 85.94\% & 13.50 & \$0.78 \\
53 & Qwen3 32B & gpt-oss-20b & 85.86\% & 6.43 & \$0.49 \\
54 & Ministral 3 8B & Qwen3 32B & 85.80\% & 9.41 & \$0.04 \\
55 & Qwen3 32B & Qwen3 32B & 84.82\% & 5.98 & \$0.62 \\
56 & Kimi K2.5 & Claude 3 Haiku & 80.41\% & 35.09 & \$0.98 \\
57 & Qwen3 32B & Claude 3 Haiku & 80.00\% & 7.86 & \$0.67 \\
58 & Qwen3 Next 80B A3B & Claude 3 Haiku & 80.00\% & 35.17 & \$0.59 \\
59 & Qwen3 Next 80B A3B & Claude Opus 4.6 & 78.00\% & 31.01 & \$2.96 \\
60 & Kimi K2.5 & Ministral 3 8B & 77.84\% & 40.79 & \$0.97 \\
61 & Kimi K2.5 & Qwen3 Next 80B A3B & 77.20\% & 37.64 & \$1.00 \\
62 & Qwen3 Next 80B A3B & Ministral 3 8B & 77.00\% & 38.55 & \$0.55 \\
63 & Qwen3 Next 80B A3B & Claude Haiku 4.5 & 76.50\% & 32.33 & \$1.21 \\
64 & Qwen3 Next 80B A3B & gpt-oss-120b & 76.50\% & 34.72 & \$0.52 \\
65 & Qwen3 Next 80B A3B & Qwen3 32B & 76.00\% & 30.64 & \$0.42 \\
66 & Qwen3 Next 80B A3B & Qwen3 Next 80B A3B & 76.00\% & 36.44 & \$0.54 \\
67 & Qwen3 Next 80B A3B & Kimi K2.5 & 75.50\% & 36.37 & \$0.79 \\
68 & Qwen3 Next 80B A3B & gpt-oss-20b & 75.00\% & 32.70 & \$0.48 \\
69 & Kimi K2.5 & gpt-oss-120b & 74.49\% & 32.23 & \$0.95 \\
70 & Kimi K2.5 & gpt-oss-20b & 74.09\% & 25.65 & \$0.77 \\
71 & Kimi K2.5 & Kimi K2.5 & 73.58\% & 44.39 & \$1.34 \\
72 & Kimi K2.5 & Claude Opus 4.6 & 73.33\% & 28.62 & \$2.79 \\
73 & Kimi K2.5 & Claude Haiku 4.5 & 73.20\% & 26.98 & \$1.36 \\
74 & Claude 3 Haiku & gpt-oss-120b & 72.19\% & 8.39 & \$0.32 \\
75 & Kimi K2.5 & Qwen3 32B & 72.16\% & 30.32 & \$0.92 \\
76 & Claude 3 Haiku & gpt-oss-20b & 71.43\% & 8.42 & \$0.32 \\
77 & Claude 3 Haiku & Qwen3 Next 80B A3B & 71.07\% & 17.12 & \$0.39 \\
78 & Claude 3 Haiku & Kimi K2.5 & 71.01\% & 14.23 & \$0.53 \\
79 & Claude 3 Haiku & Ministral 3 8B & 69.28\% & 12.40 & \$0.32 \\
80 & Claude 3 Haiku & Qwen3 32B & 59.30\% & 6.29 & \$0.29 \\
81 & Claude 3 Haiku & Claude 3 Haiku & 54.37\% & 7.28 & \$0.30 \\
\bottomrule
\end{tabular}
\end{table}

\end{document}